\documentclass[11pt]{article}

\usepackage[final]{acl}

\usepackage{times}
\usepackage{latexsym}
\usepackage{booktabs}
\usepackage{array}
\usepackage{multirow}
\usepackage{makecell}
\usepackage{multirow}
\usepackage{todonotes}
\usepackage{amsmath}
\usepackage{amsfonts}

\definecolor{datablue}{RGB}{0, 102, 204}
\usepackage[T1]{fontenc}

\usepackage[utf8]{inputenc}

\usepackage{microtype}

\usepackage{inconsolata}

\usepackage{graphicx}

\usepackage{pifont}
\usepackage{xcolor}

\title{TailLoR: Protecting Principal Components in Parameter-Efficient Continual Learning}

\author{
  Marius Dragoi \\
  Bitdefender, Romania \\
  {\small \texttt{mdragoi@bitdefender.com}} \\
  \And
  Ioana Pintilie \\
  Bitdefender, Romania \\
  {\small \texttt{ipintilie@bitdefender.com}} \\
  \And
  Alexandra Dragomir \\
  Bitdefender, Romania \\
  {\small \texttt{aledragomir@bitdefender.com}} \\
  \AND
  Antonio Barbalau \\
  Bitdefender, Romania \\
  {\small \texttt{abarbalau@bitdefender.com}} \\
  \And
  Florin Brad \\
  Bitdefender, Romania \\
  {\small \texttt{fbrad@bitdefender.com}}
}

\begin{document}
\maketitle
\begin{abstract}
Parameter-efficient finetuning methods based on spectral decomposition have enabled progress in Continual Learning. In this paper we introduce \emph{TailLoR}, which utilizes the singular bases $U$ and $V$ of the pre-trained weights as a fixed reference frame to learn a low-rank update applied to the singular value matrix. A soft spectral penalty discourages updates aligned with dominant singular directions, reducing interference while routing fine-grained adaptation into the highly flexible, long-tail spectral coordinates.
\end{abstract}

\section{Introduction}
Large Language Models (LLMs) have achieved remarkable performance across diverse reasoning and generation tasks \citep{DBLP:journals/corr/abs-2303-18223, DBLP:journals/corr/abs-2402-06196}. However, adapting these models to new domains or tasks remains computationally expensive, as full fine-tuning requires updating billions of parameters. This challenge has motivated the development of Parameter-Efficient Fine-Tuning (PEFT) methods \citep{DBLP:conf/icml/HoulsbyGJMLGAG19}, which aim to adapt large pretrained models while updating only a small subset of parameters, reducing memory and training costs.

Among PEFT approaches, Low-Rank Adaptation (LoRA) \citep{DBLP:journals/corr/abs-2106-09685} has emerged as one of the most widely adopted. Motivated by the evidence that task-specific updates often lie in a low-dimensional subspace \citep{DBLP:conf/iclr/LiFLY18, DBLP:conf/acl/AghajanyanGZ20}, LoRA freezes the pretrained weights and model updates using two trainable low-rank matrices, substantially reducing the number of trainable parameters while maintaining strong performance. 
Recent works have explored structured decompositions to further improve stability. Specifically, spectral and singular value decomposition (SVD)-based methods \citet{meng2024pissa, wang-etal-2025-milora, lingam2024svft} align trainable updates with the structure of pretrained weight matrices for more efficient tuning.


Existing low-rank adaptation methods often suffer from interference between overlapping update directions, especially when models are adapted across multiple domains or sequential tasks. This is particularly problematic in continual learning, where new knowledge must be acquired without erasing prior capabilities. Since the largest singular values of the pre-trained weight matrix encode its most critical structural representations, modifications to this subspace disproportionately degrade prior knowledge.

To mitigate this, we propose a spectral regularization scheme that selectively penalizes updates to the dominant singular components while allowing greater flexibility in the lower-rank ``tail''. Our specific contributions are as following:

\begin{itemize}
    \item We introduce \textbf{TailLoR}, a low-rank adaptation method operating over the singular values of a weight matrix, coupled with a soft regularization that steers updates toward the spectral "tail", protecting the top singular values.
    \item Different from existing continual PEFT methods \cite{das-biswas-etal-2026-ella, wang-etal-2023-orthogonal} \textbf{TailLoR} requires no access to adapters from prior tasks, crucially, enabling sequential adaptation by different users while preserving the privacy of each user's task-specific parameters and training data.
    \item We evaluate \textbf{TailLoR} on a suite of continual learning tasks, demonstrating that it matches state-of-the-art methods while increasing the stable rank of the weight matrix.
\end{itemize}

\section{Related Work}

\paragraph{Spectral LoRA variants} Leveraging the spectral properties of base weights $W$ is a key strategy in PEFT. Many SVD-based approaches partition the initial weights $W_0=U \Sigma V^\top$ into frozen and trainable components. For instance, PiSSA \citep{meng2024pissa} initializes the trainable adapter with the principal singular components to accelerate convergence, while freezing the residual. Conversely, MiLoRA \citep{wang-etal-2025-milora} freezes the principal components and initializes the adapter with the minor components
. Other approaches retain the original singular bases as a reference frame rather than statically partitioning them. SVFT \citep{lingam2024svft} adapts weights via $W=U(\Sigma+M)V^\top$, where $M$ is a learned matrix with fixed sparsity patterns. Our mechanism is closely related, but instead of fixing a sparsity pattern, we learn a low-rank update $M=AB$: $W=U(\Sigma+AB)V^\top$. We soft-regularize the update with different spectral penalties that control which couplings between $u_i$ and $v_j$ are favoured, by penalizing coupling of dominant directions and guiding the model to route new task adaptations into the underutilized, long-tail spectral coordinates. While this emphasis is similar in spirit to MiLoRA and PiSSA, these methods primarily use the SVD partitioning to decompose the original weights and initialize the adapter, while our work keeps the updates parameterized with respect to the original SVD directions throughout training.

\paragraph{Spectral Approaches for Continual Learning} Continual Learning (CL) mitigates catastrophic forgetting \citep{mccloskey1989catastrophic} across sequential tasks using rehearsal \citep{DBLP:conf/nips/Lopez-PazR17,DBLP:journals/corr/abs-1906-01076,DBLP:conf/iclr/RiemerCALRTT19}, regularization \citep{DBLP:journals/corr/KirkpatrickPRVD16}, or architecture-expansion \citep{DBLP:conf/iclr/YoonYLH18,DBLP:conf/icml/LiZWSX19,DBLP:journals/corr/RusuRDSKKPH16} strategies. Recently, Parameter-Efficient Fine-Tuning (PEFT) \citep{DBLP:conf/icml/HoulsbyGJMLGAG19}, particularly Low-Rank Adaptation (LoRA) \cite{DBLP:conf/iclr/HuSWALWWC22}, has emerged as a memory-efficient paradigm for CL, solving sequential tasks via low-parameter, task-specific adapters.

To prevent new adapters from interfering with prior knowledge, recent methods impose geometric or subspace constraints. O-LoRA \cite{wang-etal-2023-orthogonal} encourages orthogonality between successive task adapters, while ELLA \citep{das-biswas-etal-2026-ella} limits interference by restricting the specific sets of parameters that can be modified when learning a new task. Other approaches leverage spectral decompositions or activation geometries: InfLoRA \cite{DBLP:conf/cvpr/LiangL24} constructs and freeezes the matrix $B_t$ such that the update subspace for the new adapter $\Delta W_t=A_tB_t$ is aligned with the current task activations and also orthogonal to old-task gradients, while NESS \citep{pham2026learning} restricts future updates to the null space of prior activation covariances. Closest to our work is OSFT \citep{nayak2025sculpting}, a full fine-tuning method that first discovers critical prior knowledge directions corresponding to the largest singular values. During fine-tuning on new tasks, gradients are projected to the orthogonal complement of these important directions. While these methods rely on hard gradient projections or activation tracking, TailLoR achieves similar protection purely through soft spectral regularization within the efficient LoRA framework.




\section{TailLoR}
Given a pre-trained weight matrix $W \in \mathbb{R}^{d_{out} \times d_{in}}$, we extract its structural geometry via Singular Value Decomposition (SVD): $W = U \Sigma V^T$,
where $U \in \mathbb{R}^{d_{out} \times d_{out}}$ and $V \in \mathbb{R}^{d_{in} \times d_{in}}$ are orthogonal matrices containing the left and right singular vectors, and $\Sigma \in \mathbb{R}^{d_{out} \times d_{in}}$ is a rectangular diagonal matrix containing the singular values $\sigma_i$. For notational simplicity, we assume the pre-trained weight matrices are square, $W \in \mathbb{R}^{k \times k}$.

Rather than applying updates directly in the standard weight basis, we parameterize them entirely within this spectral basis. We introduce a low-rank adapter defined by the product of two matrices $B \in \mathbb{R}^{d_{out} \times r}$ and $A \in \mathbb{R}^{r \times d_{in}}$, where $r \ll \min(d_{out}, d_{in})$. The updated weight matrix during fine-tuning is constructed as:
\begin{equation}
    W' = U (\Sigma + AB) V^T
\end{equation}

Similarly with \citep{sharma2024truth, wang-etal-2025-milora}, we hypothesize that the top singular vectors (the "head") encode the foundational representations shared across tasks, while the smaller singular vectors (the "tail") represent underutilized capacity. To regularize updates $\Delta W$ during sequential fine-tuning, we construct a spatial penalty matrix $\Omega \in R^{d_{out} \times d_{in}}$ that applies distinct gradient resistance to different singular subspaces.

\subsection{The Head Penalty (Subspace Protection)}
Let $\mathbf{\sigma} \in R^k$ be the vector of non-zero singular values extracted from $\Sigma$, sorted in descending order. We first normalize the singular values by their maximum component: $
    \tilde{\mathbf{\sigma}} = \frac{\mathbf{\sigma}}{\sigma_{\max}}
$ and define the raw head penalty matrix $\mathbf{\Omega}$ such that the penalty applied to any element $(i, j)$ in the adapter update is governed by the maximum relative importance of its interacting singular components: $\mathbf{\Omega}_{i,j} = \max(\tilde{\sigma}_i, \tilde{\sigma}_j)^\gamma$, where $\gamma > 0$ is a hyperparameter controlling the severity of the penalty gradient. This formulation creates a highly concentrated regularization pressure on the top principal components, which decays as the indices move toward the spectral tail.

\paragraph{Normalization and Baselines} To validate the importance of protecting the spectral head, we compare our approach against two baselines: a \textbf{tail penalty}, which reverses the normalized singular values to intentionally penalize less significant components (isolating the effect of regularization geometry), and a \textbf{uniform penalty}, which applies a flat, unstructured regularization:
($\mathbf{\Omega}_{\text{uniform}} = \mathbf{1}_{k \times k}$).


To ensure a fair comparison, we introduce a mass normalization step. We scale the raw penalty matrix $\mathbf{P}$ so that its total mass equals $k^2$ (yielding an average element weight of $1.0$). The final normalized penalty matrix $\mathbf{\tilde{\Omega}}$ is given by:
\begin{equation}
    \mathbf{\tilde{\Omega}} = \mathbf{\Omega} \frac{k^2}{\sum_{i,j} \mathbf{\Omega}_{i,j}}
\end{equation}

During fine-tuning, the adapter matrices $A$ and $B$ are optimized with this structured regularization term added to the primary task loss. To compute the penalty, we take the square root of the $\mathbf{\tilde{\Omega}}$-weighted mean of the squared adapter updates:
\begin{equation}
    \mathcal{L}_{\text{reg}} = \lambda \sqrt{\frac{1}{k^2} \sum_{i=1}^k \sum_{j=1}^k \mathbf{\tilde{\Omega}}_{i,j} (A B)_{i,j}^2 + \epsilon}
\end{equation}
where $(A B)_{i,j}$ represents the individual elements of the adapter product (a modification over the original $\Sigma$ diagonal matrix), and $\epsilon = 10^{-12}$ is a small constant ensuring numerical stability for the gradient of the square root near zero.
\section{Evaluation and Results}
In this section, we present the evaluation setup and results on Continual Learning tasks, as well as ablations on the spectral penalty strategy and analyse the evolution of the effective rank of the weight matrix for TailLoR and ELLA.

\subsection{Continual Learning}
We evaluate TailLoR on three continual learning benchmarks: Standard CL \citep{DBLP:conf/nips/ZhangZL15}, Long Sequence \citep{DBLP:conf/iclr/RazdaibiedinaMH23}, and TRACE \citep{wang2023trace}. Across all setups, we fine-tune the \textit{query} and \textit{value} projections of a T5-large backbone for a single epoch per task with rank $r=8$. For hyperparameters, TailLoR uses a single, globally searched penalty weight $\lambda$ and exponent $\gamma$ applied statically across all tasks. In contrast, following its original protocol, ELLA's $\lambda$ coefficient is optimized independently for \textit{every} task. Despite this static tuning constraint, Table~\ref{tab:main_results_clbench} shows TailLoR (head penalty) achieves highly competitive results on the Standard CL Benchmark alongside ELLA. By matching a state-of-the-art subspace partitioning method without requiring task-specific hyperparameter tuning, TailLoR demonstrates significant robustness and efficiency. In Table~\ref{tab:main_results_trace} we show results on the TRACE benchmark with 500 samples per task on a T5-large model, with TailLoR achieving the highest overall accuracy.

\begin{table*}[!ht]
\caption{Overall Accuracy (OA) comparison on Standard CL bench (Order $1, 2, 3$) and Long Sequence bench (Order $4, 5, 6$). Results are averaged across three seeds. Top scores are bolded and second best are underlined. 
}

\vspace{-6pt}
\centering
{
\setlength{\tabcolsep}{3pt} 
\begin{tabular}{lccccccc} 
\toprule
\multirow{3}{*}{\textbf{Methods}} & \multicolumn{3}{c}{\textbf{\textcolor{datablue}{Standard CL}}} & \multicolumn{3}{c}{\textbf{\textcolor{datablue}{Long Sequence}}} & \textbf{\textcolor{datablue}{Overall}} \\
& \multicolumn{3}{c}{\textbf{\textcolor{datablue}{Benchmark (SC)}}} & \multicolumn{3}{c}{\textbf{\textcolor{datablue}{Benchmark (LS)}}} & \multirow{1}{*}{\textbf{\textcolor{datablue}{Accuracy}}} \\
& Order 1 & Order 2 & Order 3 & Order 4 & Order 5 & Order 6 & {\textbf{\textcolor{datablue}{SC + LS}}} \\
\midrule
EWC & 46.30 & 45.30 & 52.10 & 44.90 & 44.00 & 45.40 & 46.36 \\ 
IncLoRA & 59.25 & 58.69 & 69.86 & 56.83 & 56.40 & 54.32 & 59.25 \\ 
SVFT & 77.76 & 77.84 & 77.00 & 70.17 & 66.75 & 73.78 & 73.88 \\ 
MiLoRA & 66.90 & 66.67 & 70.48 & 59.73 & 57.12 & 57.27 & 63.03 \\ 
PiSSA & 72.64 & 71.14 & 70.98 & 61.84 & 62.74 & 60.57 & 66.65 \\
ELLA & 78.09 & 78.37 & 78.23 & \textbf{72.64} & 67.91 & \textbf{74.15} & \underline{74.90} \\ 
TailLoR (head) & \textbf{78.87} & \textbf{79.41} & \textbf{78.62} & \underline{71.05} & \textbf{69.04} & 72.89 & \textbf{74.98} \\ 
\bottomrule
TailLoR (tail) & \underline{78.33} & 78.22 & \underline{78.24} & 70.00 & \underline{67.97} & 72.57 & 74.15 \\ 
TailLoR (uniform) & 78.22 & \underline{78.51} & 77.62 & 69.07 & 65.96 & \underline{73.96} & 73.89 \\ 
\bottomrule
\end{tabular}
}
\label{tab:main_results_clbench}
\end{table*}

\begin{table}[!ht]
\caption{Overall Accuracy and Backward Transfer comparison on the TRACE benchmark with 500 train samples per task. Results are averaged across three seeds. 
}
\vspace{-6pt}
\centering
{
\begin{tabular}{lcc}
\toprule
\textbf{Method} 
& \makecell{\textbf{\textcolor{datablue}{Overall}}\\\textbf{\textcolor{datablue}{Accuracy}}} 
& \makecell{\textbf{\textcolor{datablue}{Backward}}\\\textbf{\textcolor{datablue}{Transfer}}} \\
\midrule
IncLoRA       & 21.68 & -24.45 \\
PiSSA         & 19.99 & -23.63 \\
SVFT          & 24.07 & \textbf{-0.26} \\
MiLoRA        & 26.13 & -13.98 \\
ELLA          & 29.40 & -10.53 \\
TailLoR (head) & \textbf{30.40} & -4.60 \\
\bottomrule
\end{tabular}
}
\label{tab:main_results_trace}
\end{table}

\subsection{Spectral Penalty Strategy Ablation}


The comparison between head, tail and uniform penalties in Table~\ref{tab:main_results_clbench} confirms the intuition in prior works: principal singular directions encode important information and protecting them from excessive task-specific updates results in more efficient continual learning. Penalizing couplings between these high singular value directions encourages task-specific adaptation through the underutilized minor singular components, which improves accuracy and significantly mitigates catastrophic forgetting. Furthermore, the underperformance of the uniform baseline suggests that treating all spectral directions equally is suboptimal compared to structurally guiding adaptation toward the spectral tail.

\subsection{Evolution of Effective Rank}

To quantify the structural changes within the weight matrices during continual learning, we track the Roy-Vetterli effective rank \citep{roy2007effective}. Given the vector of singular values $\mathbf{\sigma}$ sorted in descending order, the effective rank $R_{\text{eff}}$ is defined as the sum of the singular values bounded by the spectral norm (the maximum singular value, $\sigma_{\max}$):

\begin{equation}
    R_{\text{eff}} = \frac{\sum_{i=1}^k \sigma_i}{\sigma_{\max}} = \sum_{i=1}^k \tilde{\sigma}_i
\end{equation}

A higher $R_{\text{eff}}$ indicates a flatter spectral distribution, reflecting how task-specific updates successfully utilize the lower-rank tail capacity rather than overriding the dominant singular vectors.

\begin{figure}[t]
  \includegraphics[width=\columnwidth]{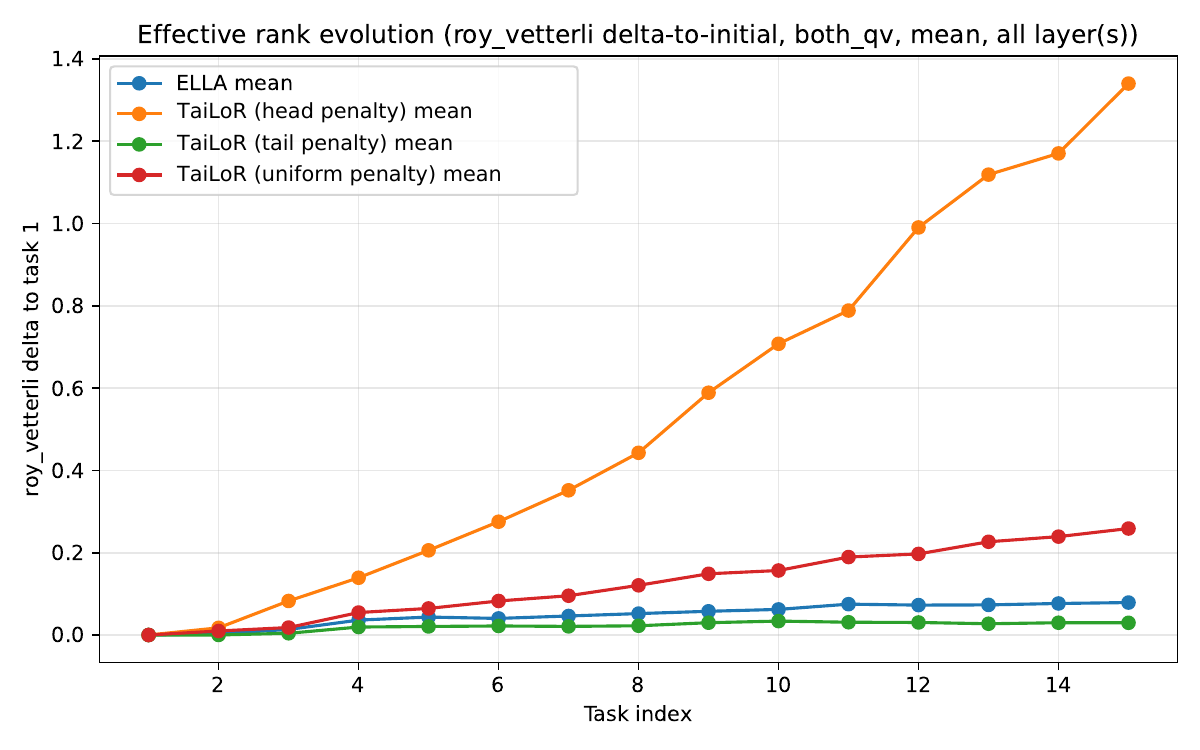}
  \caption{Effective rank analysis on Standard CL Bench (Long Order 4), relative (delta) to the initial values.}
  \label{fig:effective_rank}
\end{figure}

Figure~\ref{fig:effective_rank} illustrates the evolution of $R_{\text{eff}}$ for each regularization strategy relative to its initial value. We observe that our \textbf{head penalty} induces a consistent, significant increase in effective rank as the model sequentially learns new tasks. This empirically validates our core hypothesis: structurally penalizing modifications to the dominant singular vectors forces the network to route new task representations into the underutilized lower-rank subspace. This accumulation of information flattens the overall spectral distribution, driving up the effective rank. In contrast, methods lacking this structural awareness fail to exploit latent capacity. The \textbf{tail penalty} confines updates entirely to the already-dominant head components, resulting in a perfectly stagnant effective rank, while the \textbf{uniform penalty} yields only a marginal, unoptimized spread. Notably, while \textbf{ELLA} successfully limits interference by controlling the magnitudes of individual parameters, it ignores the underlying geometry of the weight matrix. Consequently, it leaves the spectral capacity constrained, resulting in a flat $R_{\text{eff}}$ evolution. Ultimately, these results confirm that protecting the spectral head not only preserves pre-trained knowledge but actively forces the model to exploit latent network capacity for CL.

\section{Conclusions}
In this work, we introduced TailLoR, a geometrically aware low-rank adaptation method designed to mitigate representation interference in continual learning. By selectively penalizing modifications to the dominant singular vectors, our regularization scheme successfully routes new task updates into the underutilized lower-rank tail. Empirical evaluations demonstrate that TailLoR not only preserves critical pre-trained knowledge but actively increases the effective capacity of the network, matching state-of-the-art baselines without requiring explicit weight partitioning or per-task parameter tuning.


\newpage

\section*{Limitations}
While this work demonstrates the efficacy of TailLoR on encoder-decoder architectures (T5) for continual learning, extending the spectral routing analysis to modern causal, decoder-only LLMs remains an active area of our ongoing work. Moreover, to maintain manageable computational costs, our evaluation on the TRACE benchmark utilized a 500-sample subset per task. While sufficient to demonstrate relative performance trends, future work will scale these evaluations to the full dataset.



\bibliography{custom}

\newpage

\appendix

\section{Reproducibility details}

We show in Table~\ref{tab:hyperparameter_list} the range of hyperparameters we search over for TaiLoR and the general hyperparameters for all baselines in Table~\ref{tab:params_general}. We ran all experiments on an Nvidia H200 GPU.

\label{sec:appendix_hyperparam_search}
\begin{table}[htpb]
\centering
\begin{tabular}{c p{4cm}}
\toprule
\textbf{Hyperparameter} & \textbf{Values} \\
\midrule
$\gamma$ & 
[0.5, 1.0, 2.0] \\
\midrule
$\lambda$ & 
[1e3, 2e3, 5e3, 1e4, 2e4] \\
\bottomrule
\end{tabular}
\caption{TaiLoR hyperparameters}
\label{tab:hyperparameter_list}
\end{table}

\label{sec:appendix_general}
\begin{table}[htpb]
\centering
\begin{tabular}{c p{4cm}}
\toprule
\textbf{Hyperparameter} & \textbf{Values} \\
\midrule
learning rate &  1e-3 \\
\midrule
rank &  8 \\
\midrule
epochs &  1 \\
\midrule
rank &  8 \\
\midrule
target modules &  q, v \\
\midrule
optimizer &  AdamW \\
\midrule
Weight decay & 0 \\
\midrule
LoRA dropout &  0.1 \\
\midrule
Random seeds &  [42, 43, 44] \\

\bottomrule
\end{tabular}
\caption{TaiLoR hyperparameters}
\label{tab:params_general}
\end{table}

\section{Head vs Tail penalty}
\label{app:head_tail_plots}
\begin{figure}[h]
  \includegraphics[width=\columnwidth]{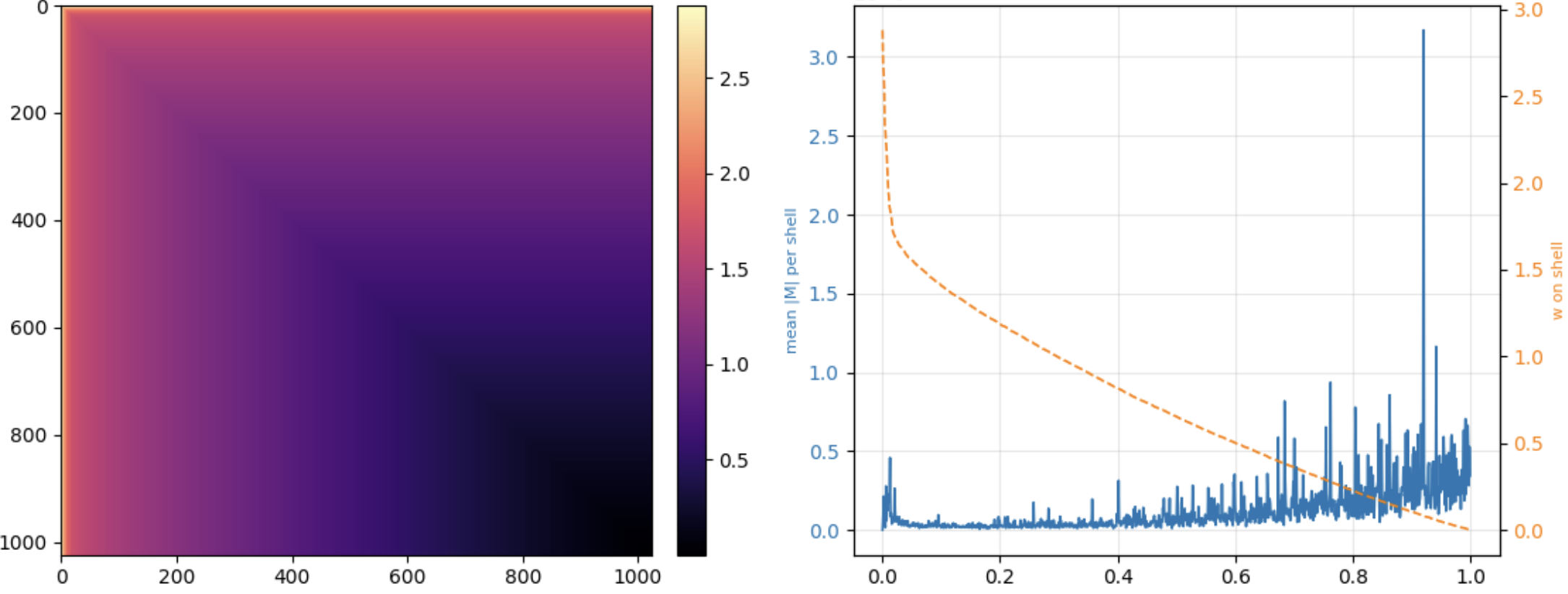}
  \caption{Head penalty matrix (left) and distribution of update magnitudes by matrix position (right)}
  \label{fig:head_penalty_matrix}
\end{figure}

\begin{figure}[h]
  \includegraphics[width=\columnwidth]{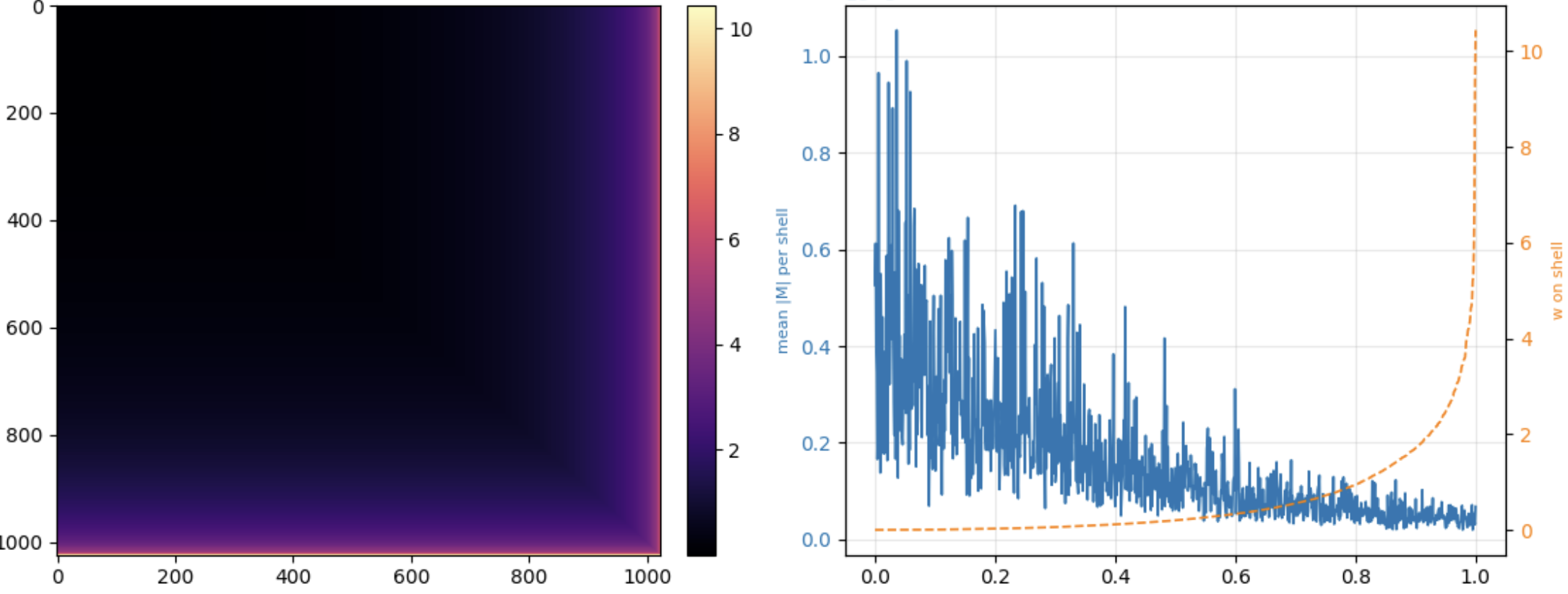}
  \caption{Tail penalty matrix (left) and distribution of update magnitudes by matrix position (right)}
  \label{fig:tail_penalty_matrix}
\end{figure}

We show the head (Figure~\ref{fig:head_penalty_matrix}) and tail (Figure~\ref{fig:tail_penalty_matrix}) penalty matrices as well as the distribution of updates across the matrix position. Shown on Long Order 4 for the T5-large model.

\newpage
\section{Datasets and Task Orders}
\label{app:datasets}

\begin{table*}[!t]
\centering
\caption{
The five classification datasets in the Standard CL Benchmark~\citep{DBLP:conf/nips/ZhangZL15}.}
\begin{tabular}{lllll}
\toprule
\textbf{Dataset Name} & \textbf{Category} & \textbf{Task} & \textbf{Domain} \\
\midrule
Yelp       & CL Benchmark & Sentiment Analysis               & Yelp Reviews             \\
Amazon     & CL Benchmark & Sentiment Analysis               & Amazon Reviews           \\
DBPedia    & CL Benchmark & Topic Classification             & Wikipedia                \\
Yahoo      & CL Benchmark & Topic Classification             & Yahoo Q\&A               \\
AG News    & CL Benchmark & Topic Classification             & News                     \\
\bottomrule
\end{tabular}
\label{tab:shortseq_datasets}
\end{table*}

\begin{table*}
\centering
\caption{
The 15 classification datasets in the Long Sequence Benchmark~\citep{DBLP:conf/iclr/RazdaibiedinaMH23}.} 
\begin{tabular}{lllll}
\toprule
\textbf{Dataset Name} & \textbf{Category} & \textbf{Task} & \textbf{Domain} \\
\midrule
Yelp       & CL Benchmark & Sentiment Analysis               & Yelp Reviews             \\
Amazon     & CL Benchmark & Sentiment Analysis               & Amazon Reviews           \\
DBPedia    & CL Benchmark & Topic Classification             & Wikipedia                \\
Yahoo      & CL Benchmark & Topic Classification             & Yahoo Q\&A               \\
AG News    & CL Benchmark & Topic Classification             & News                     \\
MNLI       & GLUE         & Natural Language Inference       & Various                  \\
QQP        & GLUE         & Paragraph Detection              & Quora                    \\
RTE        & GLUE         & Natural Language Inference       & News, Wikipedia          \\
SST-2      & GLUE         & Sentiment Analysis               & Movie Reviews            \\
WiC        & SuperGLUE    & Word Sense Disambiguation        & Lexical Databases        \\
CB         & SuperGLUE    & Natural Language Inference       & Various                  \\
COPA       & SuperGLUE    & Question and Answering           & Blogs, Encyclopedia      \\
BoolQA     & SuperGLUE    & Boolean Question and Answering   & Wikipedia                \\
MultiRC    & SuperGLUE    & Question and Answering           & Various                  \\
IMDB       & SuperGLUE    & Sentiment Analysis               & Movie Reviews            \\
\bottomrule
\end{tabular}
\label{tab:longseq_datasets}
\end{table*}

\begin{table*}[t]
\centering
\caption{
Task sequence orders for both SC and LS Benchmarks. 
}
\begin{tabular}{lcp{10cm}}
\toprule
\textbf{Benchmark} & \textbf{Order} & \textbf{Task Sequence} \\
\midrule
\multirow{4}{*}{Standard CL Benchmark} 
  & 1 & dbpedia $\rightarrow$ amazon $\rightarrow$ yahoo $\rightarrow$ ag \\
  \cmidrule{2-3}
  & 2 & dbpedia $\rightarrow$ amazon $\rightarrow$ ag $\rightarrow$ yahoo \\
  \cmidrule{2-3}
  & 3 & yahoo $\rightarrow$ amazon $\rightarrow$ ag $\rightarrow$ dbpedia \\
\midrule
\multirow{7}{*}{Long Sequence Benchmark} 
  & 4 & mnli $\rightarrow$ cb $\rightarrow$ wic $\rightarrow$ copa $\rightarrow$ qqp $\rightarrow$ boolqa $\rightarrow$ rte $\rightarrow$ imdb $\rightarrow$ yelp $\rightarrow$ amazon $\rightarrow$ sst-2 $\rightarrow$ dbpedia $\rightarrow$ ag $\rightarrow$ multirc $\rightarrow$ yahoo \\
  \cmidrule{2-3}
  & 5  & multirc $\rightarrow$ boolqa $\rightarrow$ wic $\rightarrow$ mnli $\rightarrow$ cb $\rightarrow$ copa $\rightarrow$ qqp $\rightarrow$ rte $\rightarrow$ imdb $\rightarrow$ sst-2 $\rightarrow$ dbpedia $\rightarrow$ ag $\rightarrow$ yelp $\rightarrow$ amazon $\rightarrow$ yahoo \\
  \cmidrule{2-3}
  & 6 & yelp $\rightarrow$ amazon $\rightarrow$ mnli $\rightarrow$ cb $\rightarrow$ copa $\rightarrow$ qqp $\rightarrow$ rte $\rightarrow$ imdb $\rightarrow$ sst-2 $\rightarrow$ dbpedia $\rightarrow$ ag $\rightarrow$ yahoo $\rightarrow$ multirc $\rightarrow$ boolqa $\rightarrow$ wic \\
\bottomrule
\end{tabular}
\label{tab:task_orders}
\end{table*}

In Table~\ref{tab:shortseq_datasets}, we list the original five datasets in the Standard CL benchmark (SC). Following \cite{DBLP:conf/acl/WangLJWWJCHWSZ23}, we only used the last four datasets.

In Table~\ref{tab:longseq_datasets}, we list the 15 datasets in the Long Sequence Benchmark (LS). 

We report all task orders used for our CL experiments
in Table~\ref{tab:task_orders}, which are adopted in previous works~\citep{das-biswas-etal-2026-ella}. Orders 1-3 refer to the Standard CL Benchmark, while orders 4-6 refer to the Long Sequence Benchmark.

\end{document}